\documentclass[journal]{IEEEtran}
\usepackage{amsmath} 
\usepackage{algorithm}
\usepackage{algpseudocode} 
\usepackage{array}
\usepackage[caption=false,font=normalsize,labelfont=sf,textfont=sf]{subfig}
\usepackage{textcomp}
\usepackage{stfloats}
\usepackage{url}
\usepackage{verbatim}
\usepackage{graphicx}
\usepackage{balance}
\usepackage{cite}
\usepackage{amsmath}
\usepackage{algorithm}
\usepackage{algpseudocode}
\algnewcommand\algorithmicforeach{\textbf{for each}}
\algdef{S}[FOR]{ForEach}[1]{\algorithmicforeach\ #1\ \algorithmicdo}

\begin{document}
\title{How the use of feature selection methods influences the efficiency and accuracy of complex network simulations}

\author{Katarzyna Musial, Jiaqi Wen, Andreas Gwyther-Gouriotis}

\maketitle

\begin{abstract}
Complex network systems' models are designed to perfectly emulate real-world networks through the use of simulation and link prediction. Complex network systems are defined by nodes and their connections where both have real-world features that result in a heterogeneous network in which each of the nodes has distinct characteristics. Thus, incorporating real-world features is an important component to achieve a simulation which best represents the real-world. Currently very few complex network systems implement real-world features, thus this study proposes feature selection methods which utilise unsupervised filtering techniques to rank real-world node features alongside a wrapper function to test combinations of the ranked features. The chosen method was coined FS-SNS which improved 8 out of 10 simulations of real-world networks. A consistent threshold of included features was also discovered which saw a threshold of 4 features to achieve the most accurate simulation for all networks. Through these findings the study also proposes future work and discusses how the findings can be used to further the Digital Twin and complex network system field.
\end{abstract}


\section{Introduction}
\IEEEPARstart{C}{omplex} network simulation can range from a wide array of approaches and design. The ability to model any real--world network results in the application of complex network systems to be ultimately endless. Currently, the most common use of complex network systems are in social networks, biological networks, transportation networks and chemical networks \cite{IEEEexample:Boccaletti2006complexnetworks} with each of these overarching topics having many sub topics, many of which overlap between topics, that all utilise the ability to simulate interactions of objects within the real--world.

Social network simulators (SNSs) are the focus for this experiment. SNSs are designed to simulate the internal workings of network dynamics while accurately simulating genuine networked systems. This is intended to address the issues of incomplete data brought on by data scarcity, privacy concerns, and a lack of ground truth. The simulations of these social networks can be created based on simulated data, hybrid data (constructed with both real and simulated data) or purely real--world data\cite{IEEEexample:kavak2019location,IEEEexample:musial2013social}. The goal of this research is to investigate the accuracy and efficiency of the network simulations created using purely real data. Currently there are no SNSs that implement real--world node attributes in a dynamic network, however, there are examples of static networks that utilise real--world attributes or features \cite{IEEEexample:gao2017community, IEEEexample:wang2007local, IEEEexample:lancichinetti2009benchmarks}. Using a static framework similar to previous SNSs the real--world features can be implemented and tested to see how different combinations of real--world features effect the efficiency and accuracy of the network simulation. 

Although there are static networks models that have implemented real--world attributes into their simulation, currently there is no example of a SNS that has utilised feature selection methods commonly used in machine learning to filter and rank features that are best suited for the simulation. Feature selection methods involve pre-processing data using filter, wrapper, embedded methods or hybrid methods in an attempt to select more relevant features within a data set \cite{IEEEexample:2015featureselection}. In terms of implementing a feature selection method within the SNS the methods used to select features in machine learning will be transferred into a complex network system context and implemented in order to achieve the best possible feature combination within the SNS \cite{IEEEexample:tabassum2018social}.

In this report research into feature selection methods and the data sets used for the experimentation will be presented in the background research (Section~\ref{background}). The subsequent significance, aims, research questions and objectives will then be stated, followed by a detailed analysis of the methodology as well as the experimental setup (Section~\ref{model}), followed by the results (Section~\ref{results}), future work (Section~\ref{FW}) and finally the concluding statements (Section~\ref{con}).

\section{Background}
\label{background}
Three components need to be considered prior to development of the feature selection method for the SNS: (i) SNSs as a whole and (ii) the machine learning techniques used in order to evolve the SNS to include relevant real-world features. From this research, it is understood how feature selection methods can be used within the SNS context. 
\subsection{Social Network Simulation}
Social network systems have been a relevant topic for over 40 years and were originally discussed in psychology and ecology to describe the interactions between humans and the interactions between animals in their own habitats. The most famous version of an early social network is the karate club network \cite{IEEEexample:Zachary1977nformationFlowModel} which is now commonly used as a baseline test for new link prediction or community detection algorithms \cite{IEEEexample:Hoda2015ConferenceOnComputing}. The karate club network simulates 34 members with connections between themselves constituting to a friendship involving those who frequently interacted with one another outside of club activities.

The karate club network represents a starting point when looking into SNS, however, with its lack of node heterogeneity defined as the quality or state of being diverse in character or content \cite{IEEEexample:Fletcher2007heterogeneity}, means the nodes within the network have no features to help determine connections. Heterogeneity is one of the components used by the Social DNA (sDNA) to help drive connections between nodes in various social network simulations \cite{IEEEexample:ashraf2019simulation, IEEEexample:JiaqiDTCNSsimulation2023}, thus in order to improve the accuracy of the sDNA simulations, distinguishable features between the nodes must be included to generate heterogeneity. Increased heterogeneity has been proven to improve complex network modelling of the real world such as in the case of infection rates mitigation or dispersion techniques which can be more accurately applied reducing epidemic size or even completely removing malware from a system \cite{IEEEexample:fan2021disaster, IEEEexample:Wang2021epidemicspread, IEEEexample:Jia2018heterogeneousinfection}. However, not all features improve heterogeneity, thus it is important to select the features which encapsulate the most varied and detailed representation of the objects being simulated by the nodes in the SNS. Thus, by developing feature selection methods for the SNS, a more accurate simulation of real-world networks based on heterogeneity and the sDNA can be achieved.

\subsection{Feature Selection} \label{Feature Selection}
The goal of the feature selection method is to select the most relevant node features to use in the SNS. The following background research will highlight feature selection techniques used in classification problems to provide context when applying the feature selection methods to the SNS.

\subsubsection{Filter Methods}
    Filter methods select features based on a performance measure calculated by various metrics. The filtering process depends on the search strategy which could include backward elimination, forward selection, heuristic feature subset selection and bidirectional selection. \cite{IEEEexample:2015featureselection}. Each of these searching strategies would be applicable to the features within the SNS, however, a filter method is only as useful as the evaluation techniques being used to filter out the features. Thus, the following feature evaluation methods: (i) variance, (ii) multi-collinearity, (iii) Laplacian and (iv) mutual information will be examined.
    
    Variance ranks each of the features based on how unique its information is compared to the other features in the set. A feature with a low variance means that its values are similar to another feature and thus not useful in differentiating the nodes within the system \cite{IEEEexample:Liu2007ComputationalFeatureSelection}. 

    Multicollinearity ranks features based on the correlation between any two features. In this way it is similar to basing ranks off variance, however, this method will ensure that if two features are closely correlated one of the features will be ranked lower ensuring that the more distinct feature is included within the feature groups of the simulation. An example for cars when looking at horsepower, cars tend to have a larger engine-size when the horse power is larger thus, only one of these features should be chosen \cite{IEEEexample:featureselectionsurvey2014}. 
    
    The Laplacian score is an unsupervised filter method which constructs a nearest neighbour graph to model the local structure of the feature space. The selected features are the ones that best respect the nearest neighbour structure. The Laplacian score is calculated using a diagonal matrix of the weighted connections between nodes which represent the features. The nodes obtain a connection whenever a feature is "close" enough to another feature. This is determined using the K nearest neighbour of the nodes and if two features sit within a cluster they are considered connected. Once the diagonal matrix is calculated the features with the lowest score are considered the most relevant as they best fit the local structure of the nearest neighbour structure \cite{IEEEexample:Laplacian2012}. 

    Mutual information (MI) is typically a supervised method that will be modified for an unsupervised context. It is calculated using relative entropy which measures the distance between two distributions \cite{IEEEexample:UnsupervisedFeatureSelection2012}. Using this the MI is the sum of the log of the join probability of two features over the marginal probabilities of the two features \cite{IEEEexample:featureselectionsurvey2014}. Having a MI score of 0 means there is no MI between the two features. In the context of an unsupervised filter method this is desirable. If two features have a high MI score it means having both features in the feature space is not necessary \cite{IEEEexample:Zheng2018FeatureEngineering}. 
    

    
\subsubsection{Hybrid Subset Method}
    The FS-SNS model uses a hybrid method to combine the filtering methods discussed above. Most hybrid methods use a filter method of some kind to choose promising features first, then wrapper methods are applied on the retrieved features to test best combinations of features \cite{IEEEexample:Canedo2018FeatureSelection}. There are various issues with the hybrid feature selection algorithms. The main two are that the hybrid approaches can still require a significant number of wrapper evaluations despite the fact that many unnecessary features are removed through the initial filter method. Secondly, due to the removal of features by the filter method the hybrid methods will not take into account interactions between the selected features and the pruned features. This may lead to the loss of opportunities to achieve the highest level of precision. \cite{IEEEexample:XuelianTextClassification2019}. However, with these limitations the hybrid method deployed in the FS-SNS shown in section \ref{3B} enables features to be initially ranked \cite{IEEEexample:2015featureselection} and then tested based on their rankings in order to prevent information loss from completely removing features and to show the accuracy of the SNS in correlation to the number of features included.

\section{SNS Model and Experimental Setup}
\label{model}

\subsection{Base SNS} \label{3A}
The feature selection algorithms were tested using an existing SNS structure \cite{IEEEexample:HeterogeneousJW2023, IEEEexample:ashraf2019simulation} referred to as the Base SNS in this article. The Base SNS itself is made up of 4 modules, the Feature, Network, Main and Evaluation modules. The Feature module represents the node features either through crisp representation or fuzzy representation. Crisp representation was used for this experiment which converts all the feature values to be between 0 and 1. The feature distances are then calculated for each node due to the interaction cost being replaced by the number of edges. The Network module uses the feature distance calculation with weights from the Social DNA to determine links between nodes \cite{IEEEexample:ashraf2019simulation}. The pDNA (preferential attachment) and hDNA (heterogeniety) weights are assigned to the nodes features between -1 and 1 where -1 means it is not used while 1 is the highest weight. They are then optimised using the HyperOpt function. 

The HyperOpt \cite{IEEEexample:HyperOpt2015} optimises the value space of the sDNA. Degree Distribution was chosen as the metric to optimise and evaluate the SNS based on a previous study \cite{IEEEexample:HeterogeneousJW2023} and to enable a global view of the network to be assessed as opposed to exact link predictions between nodes. The HyperOpt package uses Bayesian optimisation to combine the prior distribution of the function $f(x)$ with the posterior of the function, found using a sample of information from the data. The posterior is then used to find the maximum point of the function $f(x)$ based on a criterion \cite{IEEEexample:BayesianOptimization2019}. The hyperopt algorithm was used alongside the Tree-Structured Parzen Estimator (TPE) algorithm which searches the parameter space efficiently using a tree structure to sort the space into "good" and "bad" samples where "good" samples achieve a better result based on the optimised metric. The algorithm will continually search down the path that is more likely to produce a better result calculated using probability kernels $P(x|y)$, which represents the conditional probability of hyper parameters $x$ given that the objective function yielded a "good" result $y$ \cite{IEEEexample:TPE2023}.

Finally the evaluation of the SNS as mentioned was extended by the Social DNA study \cite{IEEEexample:HeterogeneousJW2023, IEEEexample:JiaqiDTCNSsimulation2023}. However, for the purpose of this study the focus was put on the Degree Distribution to measure the occurrences of nodes with a specific number of degrees across the entire population of the network to get a global view of how the FS methods affect the connections between nodes. The efficiency of each network simulation will also be assessed.

\subsection{Feature Selection with Social Network Simulations}\label{3B}
\begin{figure}[!t]
\centering
    \includegraphics[scale=0.35]{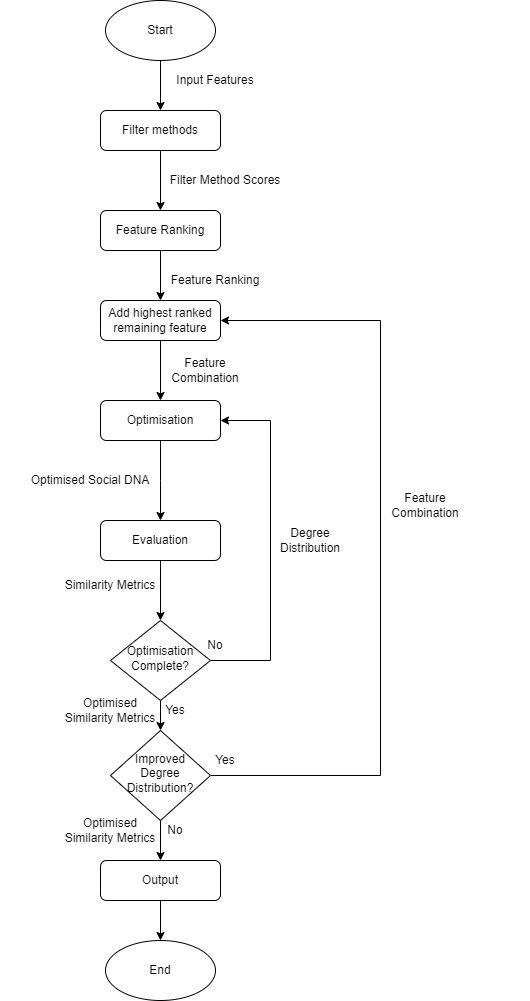}
    \label{fig:Flow diagram of FS-SNS}
    \caption{Flow diagram of the FS-SNS architecture}
\centering
\end{figure}

The following design incorporates the FS methods discussed in \ref{Feature Selection} with the Base SNS structure. The FS-SNS has two components, the first being the feature ranking followed by the feature combination wrapper. The two processes combined act as a hybrid method to test the possible feature combinations in line with their individual rankings, seen in figure \ref{fig:Flow diagram of FS-SNS}. 

To calculate the feature ranking the filtering metric scores are calculated for each feature and summed along with a weighting based on the number of times it is ranked as 1st, 2nd or 3rd amongst the various filtering metrics.

The final scores (FS) are calculated by incorporating feature rank weightings multiplied by the sum of the filtering metrics scores with $R$ and $S$ denoting feature rank and filter metric scores respectively. 

\begin{align}
FS = \sum(S \times \frac{(\sum R(1)\times3 + (\sum R(2))\times2 + (\sum R(3))}{3})
\end{align}

The feature combination wrapper loops through the feature ranking adding one feature at a time starting at the top of the ranking. Each iteration of the feature combination runs through the optimisation thus each combination is tested under one optimisation run before a new preferential attachment and homophily value is assigned to the features. The network similarity of the simulation will be measured for each feature combination inputted. The wrapper has a built in stop loss which will end the process if the error of the simulation computed using the Jensen–Shannon (JS) divergence \cite{CABRAL201473} of the degree distribution is not improved after a new feature combination is tested. The features that achieve a degree distribution error closest to 0 will be chosen as the optimal features for the SNS. This processed is outlined in the feature selection algorithm~\ref{alg:FS}

\begin{algorithm}
    \caption{Feature Selection Algorithm}
    \label{alg:FS}
    \begin{algorithmic}
        \State \textbf{Input:} Feature Ranking (FR)
        \State \textbf{Output:} Selected Feature Combination (SFC)
        \State \textit{Initialization:} current feature combination (FC)
        \State \textit{Initialization:} Global Max Degree Distribution (GMDD)
        \ForAll{$F \in \text{FR}$}
            \State \textit{Initialization:} Max Degree Distribution (MDD)
            \State $FC \gets FC \cup \{F\}$
            \State Set pDNA and hDNA
            \For{$i < \text{max\_eval}$} 
                \State Create sDNA
                \State Generate Network
                \State Evaluate Network
                \State \textit{Initialization:} Current Degree Distribution (DD)
                \If{$DD > MDD$}
                    \State $MDD \gets DD$
                \EndIf
            \EndFor
            \If{$MDD > GMDD$}
                \State $SFC \gets FC$
                \State $GMDD \gets MDD$
            \Else
                \State \textbf{Break}
            \EndIf
        \EndFor
        \State \textbf{Output:} $SFC$
    \end{algorithmic}
\end{algorithm}

\subsection{Experimental Setup} \label{3C}
The experimental setup requires data processing and cleaning, followed by the setting of parameters used for each experiment. 
\subsubsection{Data Sets Processing and Cleaning}
The following data sets are used in the experiments and with the information about their basic characteristics is shown in Table~\ref{tab5}.

\begin{table}
\begin{center}
\caption{The data sets with their file name and repository are shown in the Appendix}
\label{tab5}
    \begin{tabular}{ |p{2cm}|p{1.5cm}|p{1.5cm}|p{1.5cm}| }
        \hline
        Network & Nodes No. & Edges No. & Features No. \\
        \hline
        Ant & 116 & 4,536 & 24 \\
        Baboon & 20 & 27 & 6 \\
        Barn Swallow & 17 & 122 & 7 \\
        Twitch & 675 & 844 & 10 \\ 
        Wolf & 20 & 181 & 3 \\
        Songbird & 109 & 1,026 & 16 \\
        Bison & 26 & 222 & 4 \\
        HighTech & 21 & 145 & 10 \\
        Deezer & 3,254 & 3,257 & 7 \\
        Vampirebat & 19 & 72 & 15 \\
        \hline
\end{tabular}
\end{center}
\end{table}

The repositories used to obtain the datasets include: the Animal Social Network Repository (ASNR) \cite{IEEEexample:Sah2019multispecies}, Stanford Network Analysis Project (SNAP) \cite{IEEEexample:snapnets} and UCINET IV Datasets \cite{IEEEexample:UCINETIVDatasets}.

The goal of the preprocessing and cleaning was to ensure each of the networks follow the below criteria: 
\begin{enumerate}
    \item {Network is not bipartite}
    \item {Nodes share the same feature set}
    \item {Node features have specific meanings. (e.g. genotypes are a typical counter-example as they need to be reprocessed to be fed into the SNS)}
\end{enumerate}

These criteria were used to ensure the data sets would be suitable for the FS-SNS structure (https://github.com/AndreasGG01/FS-SNS).

For the FS-SNS, networks must be converted into a graphml for features to be extracted. For large graphs such as the Twitch and Deezer networks, a custom sampling method is required that extracts nodes from the large network ensuring that all nodes have at least one connection to another node in the sampled graph. For example, the Twitch network consists of 7,000 nodes and 35,000 edges, however, due to the limitations of the Networkx package and the SNS simulator, creating a graph of that size and simulating it was too computationally demanding, thus a smaller subset of the network was taken. For the Twitch network the number of edges were limited to 1,000. To ensure the down sampled Twitch network was not bipartite, a random node was selected from the original network and 2-5 nodes would be added to the node from its edge list. The added nodes would then go through the same process. This continued until the number of edges in the network reached the limit of 1,000. Future work could improve this sampling process for the FS-SNS as other algorithms can be used to maintain specific network structures during down sampling \cite{IEEEexample:samplingsocialnetwork}. 

Once the features have been extracted they require cleaning in order to meet the requirements mentioned above. For missing values, depending on the number of missing values for the node or feature, either the feature or node were removed. This was the case for many of the SNAP features as there were upwards of 100 features which the majority were null. For the categorical features, currently the SNS has only been designed to simulate binary categories such as sex, meaning that many categorical features needed to be encoded into binary. Finally, the network requirement of no bipartite networks, meant nodes which were detached from the main network had to be removed which occurred for the Songbird and ant network resulting in 11 nodes being removed from the songbird network and 2 from the ant network. The following table highlights the cleaning that was conducted for each network. 
\begin{table}
\begin{center}
\caption{Networks with nodes and features that were removed or encoded}
\label{tab6}
    \begin{tabular}{ |p{1cm}|p{2cm}|p{1cm}|p{1cm}| }
        \hline
        Network & No. Nodes Removed & Features Encoded & Features Removed \\
        \hline
        Ant & 2 & Group Period & NA \\
        \hline
        Baboon & NA & Dispersal & Date Feature \\
        \hline
        Wolf & NA & Sex & NA \\
        \hline
        Songbird & 11 & NA & NA \\ 
        \hline
        Bison & NA & NA & Weight \\
        \hline
        HighTech & NA & Department Partition \newline Partition Level & NA \\
        \hline
        Twitch & NA & NA & All features with null values \\
        \hline
        Deezer & 1505 & NA & All but first 7 features \\
        \hline
\end{tabular}
\end{center}
\end{table}


\subsubsection{Parameter Setting}

The encounter rate and random interference were the only parameters set for the Network module as the interaction cost parameter is not utilised when the number of edges is set. The encounter rate determines how many times each node encounters another node. The default is for the encounter rate to be set to 1 so that every node sees another node once. The random interference parameter helps identify node pairs that score the same during evaluation by adding the value to the node pari scores. The random interference must be kept small so that it doesn't significantly change the evaluation value thus a value of 0.001 was chosen. A random seed of 50 is used for the network Formation function to ensure the same network will be formulated with each test and a random see of 42 was used for the rstate in the HyperOpt optimiser to ensure that the optimised values didn't changed across the same experiment with the same features. Finall the  the evaluation function utilises two parameters, that being the numbin and the ZeroAlternative. The numbin should be the same magnitude as the number of nodes in the network thus for the majority of networks under 100 nodes a numbin of 100 was used while for networks with closer to 1000 nodes a numbin of 1000 was used. In terms of the ZeroAlternative the majority of networks used $10^{-5}$, however, the twitch network used $10^{-7}$ and the SNAP networks all used a value of $10^{-9}$. This was to ensure the distributions will be filled by a value if a certain bin does not contain any values as no node has that specific distribution. This allows for continuous similarity measures to still be used for those metrics such as shortest path distribution which would frequently not be calculated without a correct ZeroAlternative being set.

\section{Results}
\label{results}
Below results show the effects of utilising the feature selection method (FS-SNS) alongside the Base SNS discussed in Section \ref{3A}. For selected simulated networks the feature rankings (\ref{tabr2}, \ref{tabr9}) will be shown to explain the increase in accuracy of the simulated networks. Results from the Base SNS and FS-SNS are compared to highlight which filtering methods generalised over the largest number of networks to achieve the closest degree distribution similarity. Finally, the efficiency of tested methods will also be considered for each of the networks.

\begin{figure}[htp]
\centering
    \includegraphics[scale=0.5]{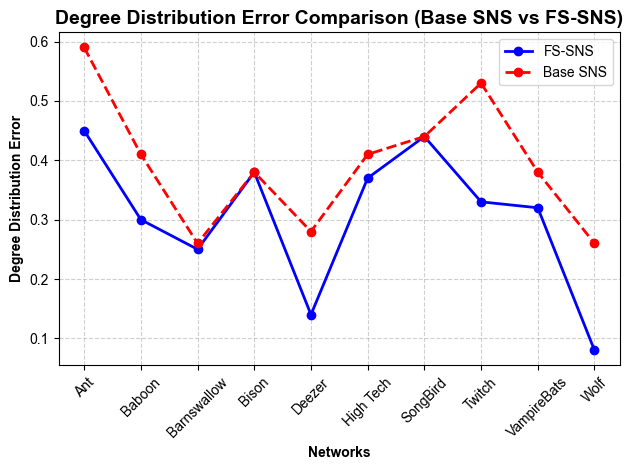}
    \caption{Base SNS vs FS-SNS based on degree distribution error (JS Divergence) across the 10 chosen networks}
    \label{Figr1}
\centering
\end{figure}

The degree distribution error comparison (Figure \ref{Figr1}) shows the comparison between Base SNS and FS-SNS when the degree distribution error score calculated by the JS divergence is considered. 

From Figure \ref{Figr1}, for 8 out of 10 data sets FS-SNS performs better than Base SNS. 
In the cases of Deezer and Wolf networks, FS-SNS performs much better than any other case when considering degree distribution similarity error. The simulated networks performed better, due to their overall structure being closely aligned with scale free networks and due to the nodes having a stronger heterogeneity than the simulated networks  which performed worse \cite{OnlineSocialNetworks2015}. For example, the Ant network falls closely in line with the small world network topology. This is a significant reason for why the Ant network performed poorly compared to other networks being simulated. The Ant node features were informative, however, the SNS models homophily and preferential attachment phenomena, which are good indicators of human social networks (such as online social media like Twitch and Deezer)~\cite{IEEEexample:ashraf2019simulation, IEEEexample:snapnets} but not small world networks. In the case of the Ant network there is no consideration of which ant is more popular, except for the queen, as all the other ants are essentially homogeneous in practise which makes their social connections much closer to a small world network. The Deezer network for example took social connections between humans from a social media networking site which was one of the best performing simulations for the FS-SNS with a degree distribution error of 0.14. In this case, the Social DNA aligns with the reasoning behind the connections formed within the social network and thus the FS-SNS was able to simulate it with greater accuracy.

{\renewcommand{\arraystretch}{1.2}
\begin{table}
\begin{center}
\caption{Base SNS to FS-SNS Results Comparison }
\label{tabr1}
    \begin{tabular}{ |p{1.30cm}|p{1.9cm}|p{0.5cm}|p{1cm}|p{0.5cm}|p{1cm}| }
    \hline
    Network& Feature Ranking Method & No. Features &FS-SNS (DD Sim) & Base SNS (DD Sim)& Acc. Increase (\%) \\
    \hline
    Ant & FR-MutualInfo & 1 & 0.45 & 0.59 & 24 \\
    \hline
    Baboon & FR-Var-Col & 1 & 0.30 & 0.41 & 27 \\
    \hline
    Barnswallow & FR-MultiCol, FR-MutualInfo & 2 & 0.25 & 0.26 & 4 \\
    \hline
    Bison & FR-Var-Col-Lap & 4 & 0.38 & 0.38 & 0 \\
    \hline
    Deezer & FR-MutualInfo & 1 & 0.14 & 0.28 & 50 \\
    \hline
    HighTech & FR-Var-Col-Lap & 1 & 0.37 & 0.41 & 10 \\
    \hline
    Songbird & FR-Var-Col-Lap & 2 & 0.44 & 0.44 & 0 \\
    \hline
    Twitch & FR-MultiCol & 1 & 0.33 & 0.53 & 38 \\ 
    \hline
    Vampirebat & FR-Col-Lap & 4 & 0.32 & 0.38 & 16 \\
    \hline
    Wolf & FR-Var & 1 & 0.08 & 0.26 & 69 \\
    \hline
\end{tabular}
\end{center}
\end{table}
}

Table \ref{tabr1} alongside Figure \ref{Figr1} highlights the overall improvement for each network between the Base SNS to the FS-SNS in terms of a percentage increase. The FS-SNS was able to improve the simulation of the 10 networks by an average of 24\% with the Twitch, Deezer and Wolf networks seeing the largest improvement from the Base SNS to the FS-SNS with over 35\%. The main reason for this could be the differences in the usefulness of network features. In all the networks mentioned the FS-SNS performed the best when one feature was selected. This highlights the fact that by using the unsupervised methods to first rank the features in terms of importance to the simulation, the simulator can get a more accurate reading on the connections each node has. The Baboon network achieves the highest result by only including the highest ranked feature from the feature selection with a degree distribution similarity error of 0.30 and once a second feature was added the degree distribution error rose to 0.46. 

Table \ref{tabr1} also shows that the most frequent number of features used to achieve the best results were a single feature. The remaining four feature combinations to achieve the best SNS accuracy were split between 2 and 4 features with only a single network utilising every feature to score the best result for the SNS, that being the Bison network.

\begin{figure}[htp]
\centering
    \includegraphics[width = 9cm]{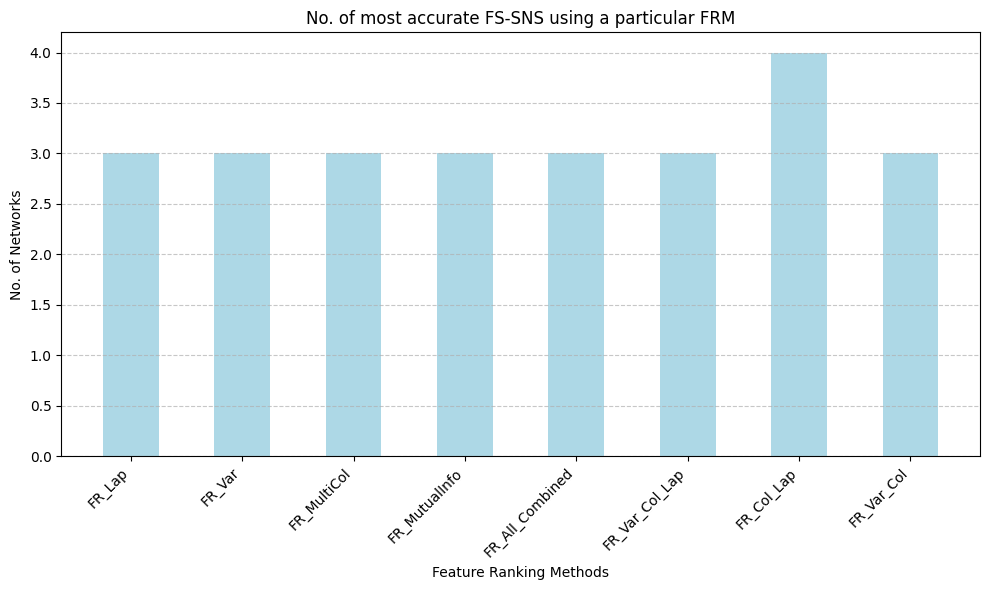}
    \caption{Base SNS vs FS-SNS based on degree distribution similarity across the 10 chosen networks}
    \label{FRM1}
\centering
\end{figure}

This highlights the importance of the unsupervised feature ranking method (FRM) that was used by the FS-SNS. The feature ranking method column in table \ref{tabr1} shows the FRM that produced the best result for each network. For networks where multiple FRMs performed the best such as for the Barn Swallow, Bison, HighTech and Songbird networks, the most frequent FRM was chosen which in most cases was the FR-Var-Col-Lap combination method. In the case of the Barnswallow network the FR-MultiCol and FR-MutualInfo methods had the frequency for the number of best performing FRM for a particular network. Ultimately the FS-SNS will be discussed as a single process, however, different FRMs should be experimented with in order to get the most optimal SNS for a given network. The best performing FRM also varies significantly with the standout FRM being the FR-Var-Col-Lap method which was the best FRM for 4 out of the 10 networks with each of the other FRMs producing the highest accuracy FS-SNS for 3 out of the 10 networks. This spread of scores across the FRMs is captured in Figure \ref{FRM1}.

The importance of utilising the right FRM to identify the most significant features in a given network can be seen in Table \ref{tabr1}. For a majority of networks only 1 particular FRM could rank the features in the most optimal order and shows that some features within the network have much higher importance for the connections between entities within the social networks. To understand why, the feature rankings of the three animal networks are shown in Table \ref{tabr2}.
\begin{table}
\begin{center}
\caption{Network Feature Rankings}
\label{tabr2}
    \begin{tabular}{ |p{2.5cm}|p{2.5cm}|p{1cm}|p{1cm}| }
    \hline
    \multicolumn{4}{|c|}{Feature Ranking FS-SNS} \\
    \hline
    Baboon Features& Ant Features& Wolf Features& FSV1 Ranking \\
    \hline
    Dominance rank at time of knockout & Nb interaction cleaners & Sex & 1\\
    \hline
    Type of dispersal & Visits to rubbishpile & Age & 2 \\
    \hline
    Group ID & Group period3 4 & Rank & 3 \\
    \hline
    Dispersal death & Nb interaction forage & - & 4 \\
    \hline
    Age in years at time of knockout & Body size & - & 5 \\ 
    \hline
    Dispersal dispersal & Visits to bood & - & 6 \\ 
    \hline
\end{tabular}
\end{center}
\end{table}

When comparing the features of the networks, the Baboon network only has one feature that provides information on the relationships between the baboons, that being the dominance ranking. As stated in the study correlating to this network, the more dominant baboons saw more connections than less dominant baboons \cite{IEEEexample:Franz2015baboonsocialnetworks}. This makes sense intuitively with our understanding of how primates interact. The other features present in the network provide little information about the interactions between baboons and instead include information about when a baboon exits the group which does not help when simulating a static network. This results in the extra features causing more noise than actual information for the SNS to draw from when simulating the relationships between the baboons. 

The opposite can be seen for the Wolf network where all three of the features provide useful information about the nodes within the graph. That is a potential reason for why the SNS performed the best on this network. However, just like the Baboon network, the Wolf network performed the best with a single feature, that being sex. The sex feature aligns well with identifying links between the primates in the wolf network as the network simulates kin relationships between primates and thus males and females would be linked together often in order to reproduce \cite{IEEEexample:UCINETIVDatasets}. The age feature would be useful to determine which of the primates are younger and thus could identify parents and children, however, there is nothing intrinsic that would highlight relationships amongst the primates. The rank feature, unlike in the Baboon network, is also not as important as it is unique for each wolf meaning that it acts more like an id than information that can be used to understand connections between the nodes. 

Finally, when looking at the Ant network the FS-SNS does improve the SNS simulation accuracy, however, this may be due to having less features to separate the ants from each other as opposed to having a standout feature to use to identify connections between ants. This is observed due to the SNS performing relatively poorly on the ant network even when only a single feature is chosen which signifies that the feature itself may not be that strong and instead is due to there being less noise in the simulation overall from the features. Based on the study of the ant social network the researchers broke the ants workers in to three groups, nurses, foragers and cleaners and found that interaction within colonies was primarily mediated by age-induced changes in the spatial location of workers \cite{IEEEexample:antnetworkscienceexpress}. This highlights a potential reason for why the features highlighting the interactions between groups resulted in the highest ranked features. The number of interactions with cleaners was the highest ranked feature with interactions with forage being 4th. This makes sense in the context of the study as interactions with particular groups would result in more connections between the workers. In the study it is also mentioned that as ants age they move through the working ranks of nurses, cleaners and forages, with no consistent size difference between the groups. This supports the interaction with cleaners as being the most prominent feature for simulation as it is the middle of the three roles amongst the working ants and thus has the most crossover with the other two groups forming the best individual feature to simulate the interactions between ants. 

The large social networks also perform better with only a single feature, however, due to the way the SNAP repository normalises its features and removes the actual feature names from the network dictionary it is hard to confirm what the features actually inform about the nodes within the networks \cite{IEEEexample:snapnets}. Thus they cannot be discussed in terms of whether having a single feature performed the best because it held useful information about the nodes or whether by only having one feature included in network the noise from other features was minimised.

The next set of networks saw little improvement from the Base SNS compared to the feature selection methods, these include the Bison, Barn Swallow and Songbird networks. The Barnswallow network saw little improvement between the Base SNS and FS-SNS even when only two features were selected, while the Bison network performed the best for both the Base SNS and FS-SNS when all features were selected and the Songbird network had no change when a single feature or all features were selected. When looking at these features in Table \ref{tabr9} it is clear to see that for the Bison and Barn Swallow networks the majority of features provide useful information about the nodes themselves with considerably less uninformative features.   

\begin{table}
\begin{center}
\caption{Feature Ranking FS-SNS for subset of network dataset}
\label{tabr9}
    \begin{tabular}{ |p{1.5cm}|p{1cm}|p{3cm}|p{1cm}|}
    \hline
    Barn Swallow Features& Bison Features& Songbird Features & FS-SNS Ranking \\
    \hline
    Sex  & Cows bred & Number Days with RFID Data & 1\\
    Mean Wing  & Age & Weighted Degree & 2 \\
    Mass & Total losses & Eigenvector Centrality & 3 \\
    StressCort & Total wins & Number of Captures & 4 \\
    MicrobialDiv  & - &  Instances of Aggression per Day &  5 \\ 
    MeanTS  & - &  Feeders Visited per Day &  6 \\ 
    Color & - & Feeding Bouts per Day & 7\\
    - & - & Num Unique Indivs w which Interacted Aggressively & 8\\
    - & - & Average Adjusted Group Size & 9\\
    - & - & Time on Feeders per Day in Minutes & 10\\
    - & - & Relative Dominance &  11\\
    - & - & Infected yes & 12\\
    - & - & Infected no & 13\\
    - & - & Male & 14\\
    - & - & Female & 15\\
    - & - & Undefined & 16\\
    \hline
\end{tabular}
\end{center}
\end{table}

 The Barn Swallow features each provide information on the bird's characteristics which intuitively would provide information to the SNS on whether two Barn Swallows have a relationship. It also makes sense for the sex of the Barn Swallows to be the highest ranked feature as male and female Barn Swallows are more likely to have a relationship for mating \cite{IEEEexample:BarnswallowASNR2016}. When looking at the Bison network it is a similar case as the network relationships are based on those Bison that dominated each other thus having information about the number of cows bred, wins and losses against other bison and their age all helps with identifying the most dominant bison \cite{IEEEexample:BisonASNR1979}. None of the features explicitly help with identifying which bison dominated another particular bison, thus it would be hard for the SNS to determine exact relationships purely based on these features. 
 
 The Songbird network shows the spread of a contagion, however, a lot of the features provide information about the environment of the birds as opposed to features of the bird such as the number of aggressive interactions the bird had, the number of visits, the time of the day, group size, etc \cite{IEEEexample:SongBird1987}. There is also a lot of information about the statistics of the network such as the weighted degree and eigenvector centrality of a node which ranked higher than other features. This information may be useful when using a topology-based simulator. However, the SNS creates global measures based on node to node interactions and similarity and there is little information within the network that helps identify characteristics of the birds apart from relative dominance and gender, which may have resulted in a more accurate SNS if ranked higher. Ultimately since none of the features highlights the spread of the contagion amongst the Songbird there was no difference in accuracy between the Base SNS and FS-SNS. 

Overall, the FS-SNS simulates networks accurately when there are both features which highlight useful characteristics about the nodes within the network such as for the Baboon, Barnswallow and Wolf network, and a network structure that is highly determined by preferential attachment and heterogeneity such as with the social media networks. Combining these two network characteristics the FS-SNS is able to achieve highly accurate simulations of a given network.

\subsubsection{Degree Distribution Comparison}
\begin{figure}[!t]
\centering
    \includegraphics[width=9cm]{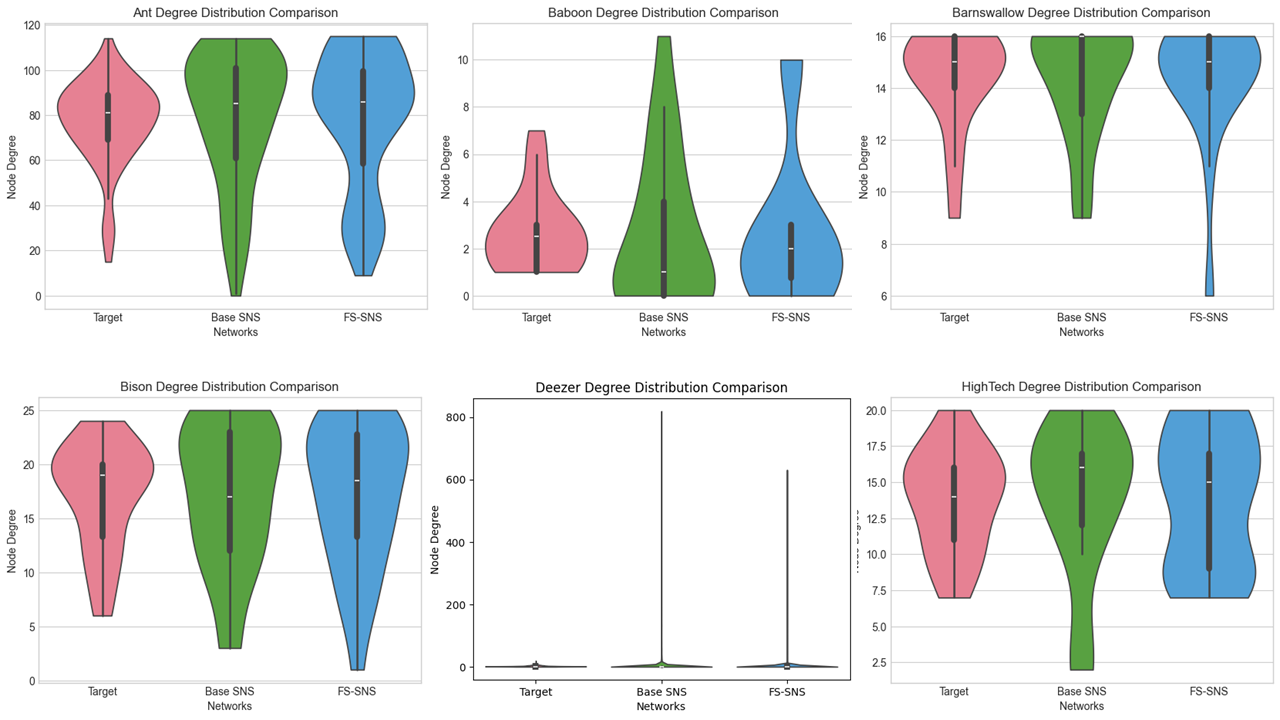}
    \includegraphics[width=9cm]{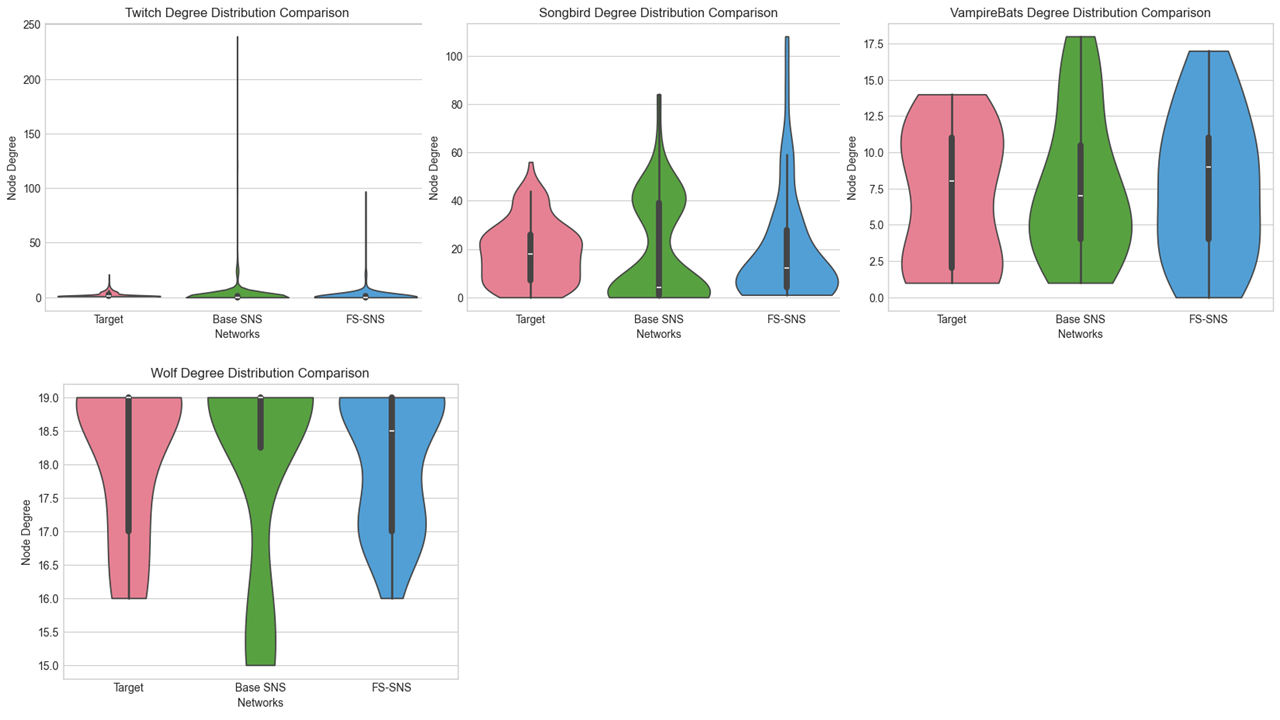}
    \caption{Degree distribution of each network comparing the target graph to the Base SNS and FS-SNS}
\label{fig3}
\end{figure}

To understand how the degree distribution similarity is measured and the differences in performance between each of the simulations using the Base and FS-SNS, the following violin plots in Figure \ref{fig3} are presented to show the degree distributions of the target network against each of the simulations.

\noindent From each of the graphs it is clear that the FS-SNS was able to match the degree distribution more closely than the Base SNS Such as the Wolf and Baboon networks, the Twitch network and the HighTech network. There are also cases where although it seems that the Base SNS has a closer distribution to the target network than the FS-SNS, the FS-SNS is still more accurate. For example the Barn Swallow network seems to have a more accurate Base SNS, however, when looking at the median line and the thick black bar which represents the interquartile range and the thin black bar which visualises the degree distribution without the outliers the FS-SNS is significantly closer to the target. In some cases it is clear that simulated networks were not able to accurately match the target distribution, such as for the Songbird network, which was seemingly the most inaccurate. For the HighTech and Bison networks the SNS was able to capture the overall shape, however, did not capture where the bulk of the degree distribution was in the target degree distributions. Similarly the majority of simulated cases have longer tails than the original target. This is due to the high encounter rate, since each node can encounter every other node once, the simulations are more likely to have outlier nodes with higher degrees than the target. This is particularly true for the large social media networks which have significantly longer tails than the target degree distribution. For other networks the long tail is for an opposite reason. Due to the simulator trying to match the degree distributions of nodes with very large degrees as well as nodes with a smaller number of degrees, but not 0. The simulator would skew a large portion of the nodes to having no degree at all which is inaccurate as all target networks are required to have all nodes connected to the graph by at least one edge. Thus, in future work it may be worth adding logic to the simulator so that, if a node does not have any edges, an additional random connection could be introduced. This issue occurs for networks such as the Wolf, Barn Swallow, HighTech and Bison network. Typically when the majority of node degrees are at the high end of the distribution the simulators will have outliers around node degree 0 while for the networks with a majority of nodes with lower degrees the outliers will have higher degrees at the end of the distribution. 

The Deezer and Twitch networks both have unique looking degree distributions due to the overall size of the network and the sparsity of connections. This is reflective of real life, as users in a large social media network have significantly fewer connections than the number of users on the platform. However, it could also be due to the sampling method discussed in Section~\ref{3C}, which aimed at reducing the number of nodes and connections used for the SNS, which also meant that a lot of nodes added had only a single other connection. In order to get a better simulation of these large graphs, improvements to the computational efficiency of software such as NetworkX and the SNS will have to be made in order to simulate the entire large network, which would be the only way to ensure the degree distribution is perfectly representative of the original large scale network. Further discussions about this limitation is in Section~\ref{FW}. 

When looking at the overall accuracy of the simulation's degree distributions, it is clear that the majority of simulations were able to get an accurate distribution which supports the idea that the SNS can simulate real-world networks when real node features are included into the social DNA. Although some networks were more accurately simulated than others. In a majority of cases the most common node degree of the distributions can be seen to line up, with the overall shape accurately modelled along with some outliers. To get a precise look at the exact score of the degree distribution accuracy of each of the SNS methods refer to Table \ref{tabr1}. From Table \ref{tabr1} the FS-SNS scored a lower degree distribution error for 8 out of 10 of the tested networks. For the two remaining networks, Bison and Songbird, the SNS accuracy did not improve from the Base SNS. The overall accuracy of the FS-SNS can be compared to a previous study which used the Jensen–Shannon divergence to measure the similarity of a simulated network's degree distribution against the target network. The majority of simulated networks from the study scored an error of 0.6 - 0.4  with the most accurate simulation achieving a score of 0.26 \cite{IEEEexample:JiaqiDTCNSsimulation2023}. These results were achieved using simulated features, however, they align well with the results achieved by the simulations of the 10 real-world networks with the best score achieved from this study being 0.08 for the Wolf network by the FS-SNS.

\subsubsection{Efficiency}
\begin{figure}[!t]
\centering
    \includegraphics[width=9cm]{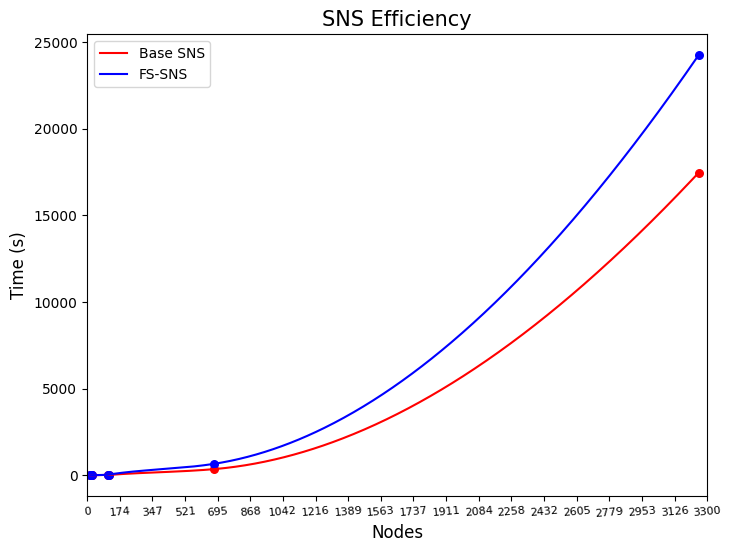}
    \caption{The efficiency of each of the feature selection methods}
\label{fig6}
\end{figure}

In terms of the computational efficiency of the different SNS, Figure \ref{fig6} compares the efficiency across the different networks.

Figure \ref{fig6} shows the difference in efficiency between the Base SNS and FS-SNS in terms of time difference needed to complete an SNS simulation for each network using the particular method. The bottom curve in red is the Base SNS, which is the most efficient, as it only runs the optimisation algorithm once and does not make any calculations to determine which features are the most relevant. The FS-SNS (red line) is only slightly more inefficient than the base SNS. This is due to the fact that the FS-SNS must run the optimisation at least two times to see if there is an improvement with the next feature combination. If the FS-SNS accuracy continues to improve with each new added feature, then there is a potential for the FS-SNS to run as many times as there are features in the network as seen for the Bison network, which would have been significantly more efficient to purely run the Base SNS. Ultimately it is due to the FS-SNS testing different feature combinations which results in the FS-SNS outperforming the Base SNS in 8 out of 10 cases. Which means that although the FS-SNS is slightly less efficient than the Base SNS the increase in accuracy demonstrated in Figure \ref{Figr1} makes the loss in efficiency worth the use of the FS-SNS. Table \ref{taber1} shows the exact time it took to run each of the feature selection methods for each network.

\begin{table}
\begin{center}
\caption{Efficiency of the Base SNS and FS-SNS for each network}
\label{taber1}
    \begin{tabular}{ |p{2cm}|p{1.5cm}|p{1.5cm}| }
    \hline
    Network& FS-SNS (s) & Base SNS (s)\\
    \hline
    Ant & 45.57 & 22.07 \\
    Baboon & 3.74 & 2.59 \\
    Barn Swallow & 5.09 & 2.8\\
    Bison & 9.12 & 5.83 \\
    Deezer & 24232.7& 17449.03 \\
    HighTech & 4 & 2.26 \\
    Songbird & 30.2 & 14.71 \\
    Twitch & 661.81 & 356.3 \\ 
    Vampirebat & 9.11 & 3.84 \\
    Wolf & 3.88 & 3.59 \\
    \hline
\end{tabular}
\end{center}
\end{table}

\subsubsection{Summary and discussion}

The results show that the inclusion of the feature selection method greatly improves the accuracy of the SNS for 8 out of the 10 tested networks. The greatest improvement found using the feature selection methods include the Degree Distribution error decreasing by 69\% with many simulations decreasing by 20\% or more. This material improvement occurred purely through selecting the highest ranked feature to run the SNS based on the feature ranking system described in Section \ref{3B}. Out of the different filtering metrics tested the FS\_Col\_Lap performed the best which utilised the Multicollinearity and Laplacian feature ranking methods. The other feature ranking methods also enabled the most accurate SNS to be achieved for 3 out of 10 networks, thus there was no way to ultimately decide on a single feature ranking method and were all considered for the final FS-SNS. In some cases, if the node features provided little information to help determine the relationships between the nodes a single top ranked feature was enough to achieve the most accurate SNS. However, in some cases a network would have multiple useful features or require the combination of multiple features to get the desired information about the node connections. In this case the FS-SNS would not be required as the most optimal SNS would be those that include all the network features. Thus, the takeaway from the analysis is that the FS-SNS should be used over the Base SNS for simulating real-world networks and that the most accurate SNS can be achieved for all networks with 4 or less real-world features from the network. 

As seen in Figure \ref{fig6} and Table \ref{tabr1}, the computational expenditure increases when more feature combinations are optimised. Ultimately the FS-SNS does increase the time the SNS takes to run in comparison to the Base SNS, however, due to the material improvements seen for the networks such as the Wolf, Baboon and Deezer network the added time is ultimately worth spending. 

Finally, the degree distributions of the target and simulated graphs were compared to highlight the overall accuracy of the SNS and through comparison with the previous study utilising the same SNS architecture the accuracy of the simulations align with those achieved in the previous study which utilised simulated features. In some cases, the FS-SNS were able to achieve more accurate results than those seen in the previous study, however, a direct comparison cannot be made due to different networks being simulated. Although it does provide validation that the accuracy achieved by the methods discussed in this article are forwarding the research within the Digital Twin and Complex Network Systems field.

\section{Future Work}\label{FW}

To extend the research conducted in this study, different values for metrics specified in the experimental setup can be tested in order to potentially improve the simulation results of the real world networks including more iterations of the HyperOpt function. The encounter rate in particular could be a focus point for further research as having an encounter rate of 1 can result in unrealistic degrees as seen in the large social media networks where it is unlikely that one user will come across every other user in the network. A change in the similarity metrics could also be tested such as measuring the clustering coefficient instead of the degree distribution which was chosen to align with previous studies using the social DNA simulator \cite{IEEEexample:JiaqiDTCNSsimulation2023}. 

Further testing of different filtering metrics to improve the feature selection model could be a point of interest for future work as well as using a different wrapping method which could tie into improving the overall structure of the SNS simulator to allow it to handle string values instead of purely categorical binary data which prevented the use of features in specific networks \cite{IEEEexample:socPokec2012}. Furthermore, designing the SNS to be able to handle bi-partite networks would allow for more networks to be tested and currently with the limitation of NetworkX and the current speed of optimisation of the Social DNA using HyperOpt very large networks ($>$10,000 nodes) as a whole are unable to be included in the experimentation. 

From the results of this study the feature selection method can be implemented in future studies to help researchers decide on features to include in their CNS and by utilising the FRM discussed in the Section \ref{results} a systematic way of testing which features to incorporate in a CNS can now be conducted for future work. Outside of node features, using the feature selection methods could improve research results when implementing topological and edge features for new DT and CNS designs.

\section{Conclusion}\label{con}

This study showcases the improvements achieved by implementing a feature selection method for the proposed SNS in the previous study \cite{IEEEexample:JiaqiDTCNSsimulation2023}. The feature selection method FS-SNS improved the accuracy of the Base SNS for 8 out of 10 network simulations. 

The results and discussion detailed the different causes in simulation accuracy and showed why implementing the feature selection method alongside the SNS improved the accuracy of the simulation. To summarise, the main influencing factor in the way the FS-SNS performed was how relevant a particular feature was to the relationships in the network. As discussed some of the networks contained little node attribute information to help determine the specific connections found in the network and thus performed worse than other simulations. Networks such as the Songbird and Vampire bat did not have enough relevant features to simulate the relationships in the graph unlike the Wolf and Barnswallow networks. Another major factor to the performance of the FS-SNS was the topology of the networks such as the Ant social network which acted similarly to a small world network where preferential attachment and heterogeneity were less important, while networks such as the social media network Deezer or dominance-based Baboon network followed a more traditional scale free architecture. These differences in network architecture were reflected in how the SNS performed. These cases show that although selecting the right real-world node attributes improves the accuracy of the SNS, to further model the networks seen in the real-world with greater accuracy, topological features, network structure and the type of connections should also be considered. With these findings and the proposed FS-SNS, researchers can continue designing simulations of real-world networks with the inclusion of real-world node features for the purpose of further advancing the research fields of Digital Twins and Complex network systems.

\section{Acknowledgements}
This work was supported by the Australian Research Council, Dynamics and Control of Complex Social Networks under Grant DP190101087.


\begin{IEEEbiography}[{\includegraphics[width=1in,height=1.25in,clip,keepaspectratio]{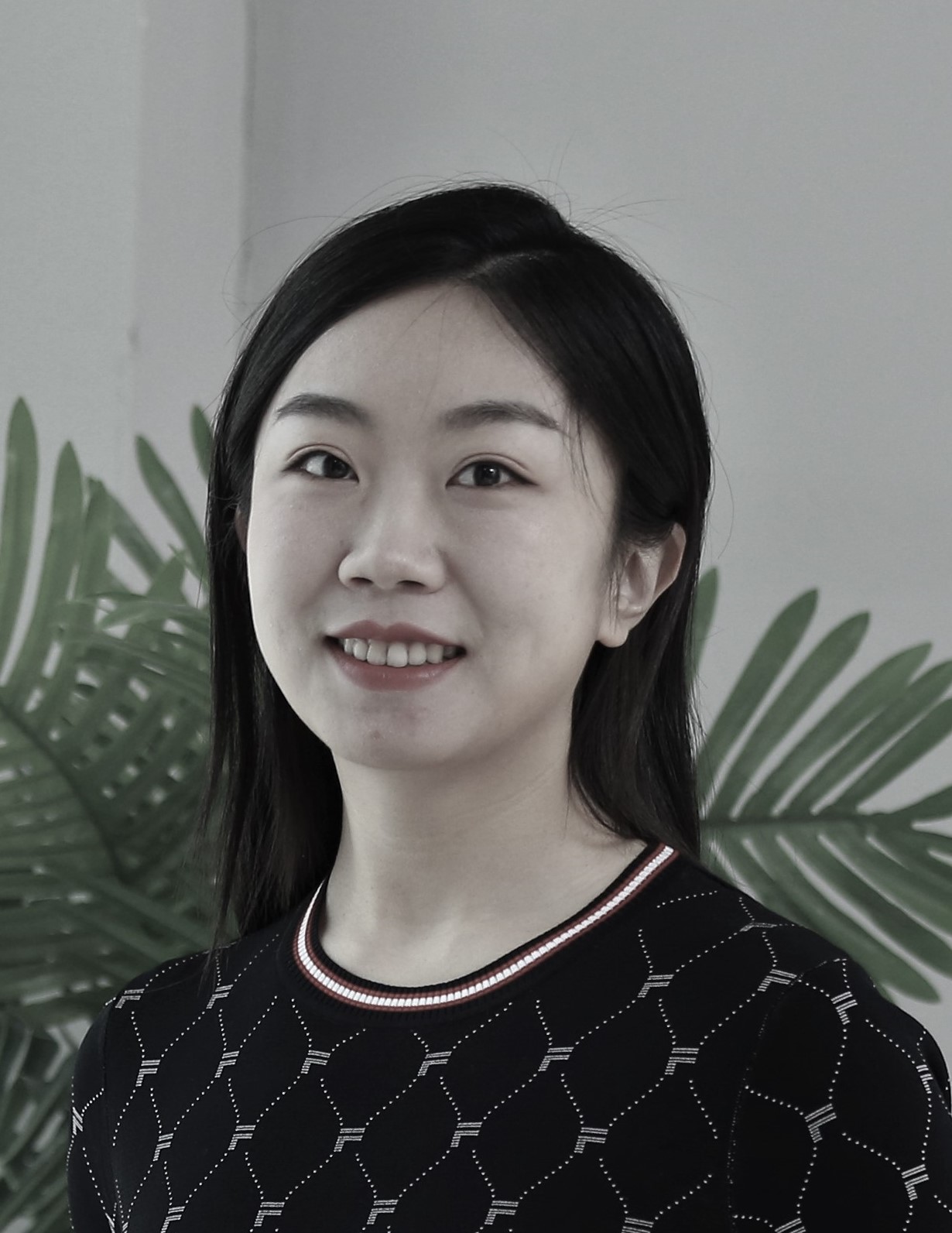}}]
{Jiaqi Wen} received the PhD. degree in Computer Science from the University of Technology Sydney (UTS), Sydney, Australia, in 2024. She is currently working as a research associate with the University of Technology Sydney. 

She worked as a research assistant in the school of economics and management, Beihang University, Beijing, China. Her research interest includes modelling complex networked systems in the Digital Twin space.
\end{IEEEbiography}

\begin{IEEEbiography}[{\includegraphics[width=1in,height=1.25in,clip,keepaspectratio]{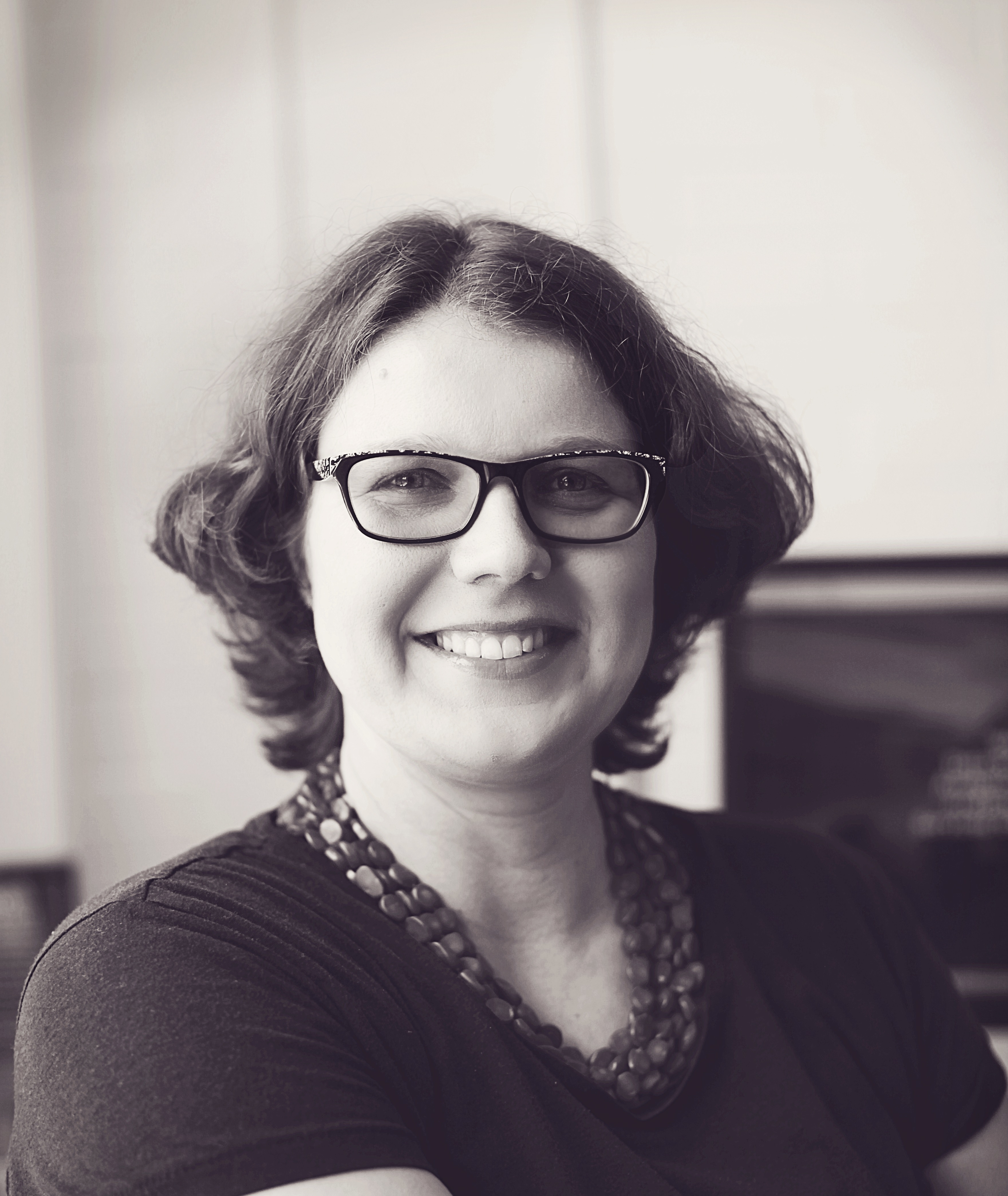}}]{Katarzyna Musial}
was awarded her PhD in Computer Science in November 2009 from Wroclaw University of Science and Technology (Poland), and in 2010 she was appointed a Lecturer in Informatics at Bournemouth University, UK. She joined King’s College London in 2011 as a Lecturer in Computer Science. In 2015 she returned to Bournemouth where she was an Associate Professor in Computing. From 2017 she works as a Professor in Network Science in the School of Computer Science and a Co-Director of the Complex Adaptive Systems Lab at the University Technology Sydney. More details can be found at: \url{http://katarzyna-musial.com}
\end{IEEEbiography}

\begin{IEEEbiography}[{\includegraphics[width=1in,height=1.25in,clip,keepaspectratio]{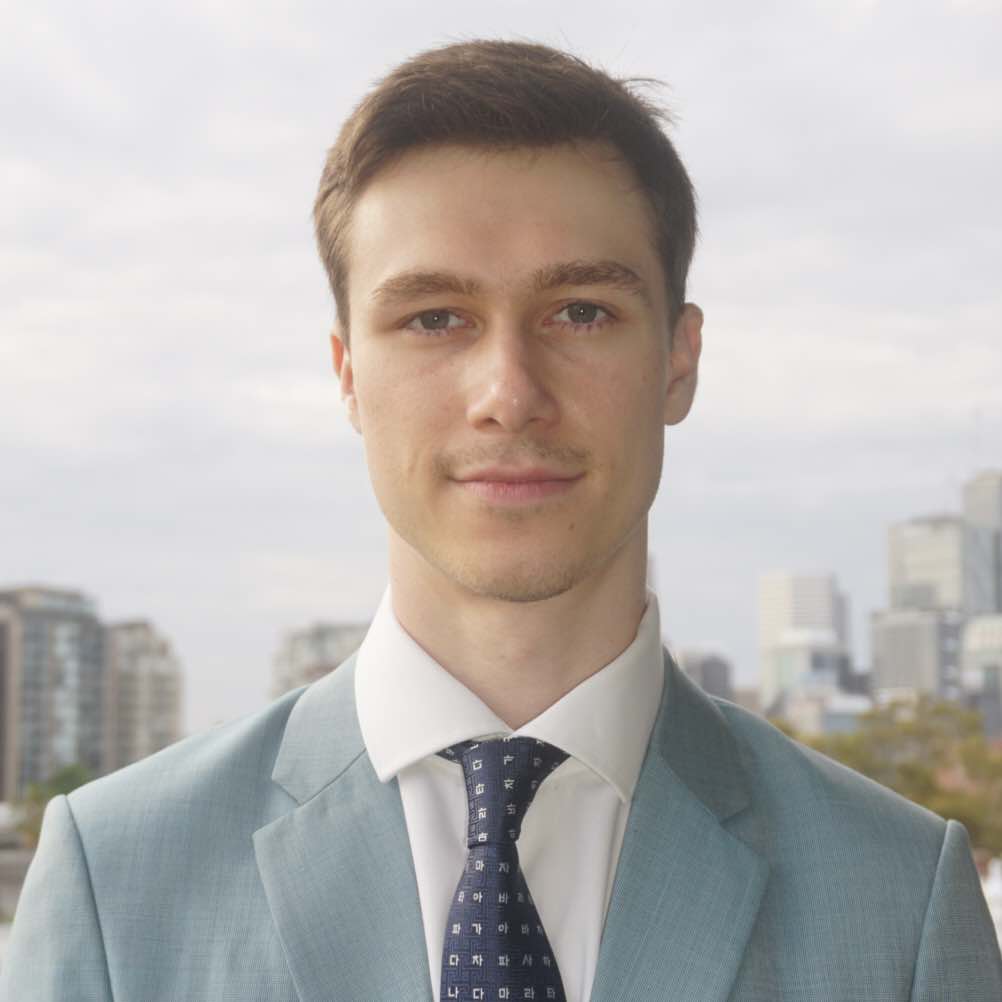}}]{Andreas Gwyther-Gouriotis}
completed his Bachelor's in Computer Science with first class Honours in November 2023 from the University of Technology Sydney (UTS), Australia. He is currently working in the field of Artificial Intelligence for a Sydney based startup, iSEAO.

He has worked for UTS as a research intern supervised by Professor Katarzyna Musial. His research interests include modelling social networks in the Digital Twin space.

\end{IEEEbiography}

\end{document}